\documentclass{article}

\usepackage[preprint]{kobi}
\usepackage{graphicx}
\usepackage{caption}
\usepackage{float}
\usepackage{lmodern}
\usepackage[utf8]{inputenc} 
\usepackage[T1]{fontenc} 
\usepackage{xcolor}
\definecolor{darkblue}{rgb}{0.1, 0.2, 0.75}
\usepackage[hidelinks]{hyperref}
\hypersetup{
    colorlinks=true,
    linkcolor=darkblue, 
    citecolor=darkblue, 
    urlcolor=darkblue   
}
\usepackage{url}           
\usepackage{booktabs}       
\usepackage{amsfonts}       
\usepackage{nicefrac}      
\usepackage{microtype}      
\usepackage{xcolor}         
\usepackage{lineno}
\usepackage{tabularx}
\usepackage{makecell}
\usepackage{threeparttable}
\usepackage{amsmath}
\usepackage{footnote}

\title{The Levers of Political Persuasion with Conversational AI}

\author{
  Kobi Hackenburg$^{1,2}$\thanks{Lead authors contributed equally. Correspondence to: \href{mailto:kobi.hackenburg@oii.ox.ac.uk}{\texttt{kobi.hackenburg@oii.ox.ac.uk}} or \href{mailto:b.tappin@lse.ac.uk}{\texttt{b.tappin@lse.ac.uk}}. $^{\dagger}$Co-senior authors.}   ,
  Ben M. Tappin$^{3}$\footnotemark[1] ,
  Luke Hewitt$^{4}$,
  Ed Saunders$^{1}$, \\
  \textbf{Sid Black}$^{1}$, 
  \textbf{Hause Lin}$^{5}$,
  \textbf{Catherine Fist}$^{1}$,
  \textbf{Helen Margetts}$^{2}$,\\
  \textbf{David G. Rand}$^{5\dagger}$
  \textbf{\& Christopher Summerfield}$^{1,2\dagger}$ \and 
  \\
{$^1$UK AI Security Institute}\\ 
{$^2$University of Oxford}\\ 
{$^3$The London School of Economics and Political Science}\\ 
{$^4$Stanford University}\\
{$^5$Massachusetts Institute of Technology}
}

\begin{document}

\maketitle
\vskip 0.2in
\begin{abstract}
\leftskip=.25in
\rightskip=.25in

There are widespread fears that conversational AI could soon exert unprecedented influence over human beliefs. Here, in three large-scale experiments (N=76,977), we deployed 19 LLMs—including some post-trained explicitly for persuasion—to evaluate their persuasiveness on 707 political issues. We then checked the factual accuracy of 466,769 resulting LLM claims. Contrary to popular concerns, we show that the persuasive power of current and near-future AI is likely to stem more from post-training and prompting methods—which boosted persuasiveness by as much as 51\% and 27\% respectively—than from personalization or increasing model scale. We further show that these methods increased persuasion by exploiting LLMs' unique ability to rapidly access and strategically deploy information and that, strikingly, where they increased AI persuasiveness they also systematically decreased factual accuracy.

\vskip 0.4in

\end{abstract}

\section*{Introduction}
\label{sec:intro}

Academics, policymakers and technologists fear that AI may soon be capable of exerting substantial persuasive influence over people \cite{lucianoHypersuasionAIsPersuasive2024, burtellArtificialInfluenceAnalysis2023, jonesLiesDamnedLies2024, rogiersPersuasionLargeLanguage2024, El-Sayed2024, Grace2024, nostaAIsSuperhumanPersuasion, bengioManagingExtremeAI2024, hsuDisinformationResearchersRaise2023, Goldstein2023, Mackenzie2023, Ipsos2023, durmus2024persuasion}. Large language models (LLMs) can now engage in sophisticated interactive dialogue, enabling a powerful mode of human-to-human persuasion \cite{Brookman2016, kalla2022personalizing, KALLA_BROOCKMAN_2020} to be deployed at unprecedented scale. However, the extent to which this will impact society is unknown. We do not know how persuasive AI models can be, what techniques increase their persuasiveness, and what strategies they might use to persuade people. For example, as compute resources continue to grow, models could become ever more persuasive, mirroring the ‘scaling laws’ observed for other capabilities. Alternatively, specific choices made during model training, such as the use of highly curated datasets, tailored instructions, or user personalization might be the key enablers of ever greater persuasiveness. Here, we set out to understand what makes conversational AI persuasive and to define the horizon of its persuasive capability.


To do so, we examine three fundamental research questions related to distinct risks. First, if the persuasiveness of conversational AI models increases at a rapid pace as models grow larger and more sophisticated, this could confer a substantial persuasive advantage to powerful actors who are best able to control or otherwise access the largest models, further concentrating their power. Thus, we ask: are larger models more persuasive? (RQ1). Second, because LLM performance in specific domains can be optimized by targeted post-training techniques, as has been done in the context of general reasoning or mathematics \cite{ouyang2022training, wei2022chain, lewkowycz2022solving}, even small open-source models—many deployable on a laptop—could potentially be converted into highly persuasive agents. This could broaden the range of actors able to effectively deploy AI to persuasive ends, benefiting those who wish to perpetrate Coordinated Inauthentic Behavior for ideological or financial gain, foment political unrest among geopolitical adversaries, or destabilize information ecosystems \cite{Goldstein2023, Goldstein2024, Wack2025}. Thus, we ask: to what extent can targeted post-training increase AI persuasiveness? (RQ2). Third, LLMs deployed to influence human beliefs could do so by leveraging a range of potentially harmful strategies, such as exploiting individual-level data for personalization \cite{Salvi2024, matzPotentialGenerativeAI2024, rogiersPersuasionLargeLanguage2024, simchonPersuasiveEffectsPolitical2024, Hackenburg2024} or by using false or misleading information \cite{jonesLiesDamnedLies2024}, with malign consequences for public discourse, trust and privacy. Thus, we ask: what strategies underpin successful AI persuasion? (RQ3).

We answer these questions using three large-scale survey experiments, across which 76,977 participants engaged in conversation with one of 19 open- and closed-source LLMs that had been instructed to persuade them on one issue from a politically balanced set of 707 issues. The sample of LLMs in our experiments spans more than four orders of magnitude in model scale and includes several of the most advanced (``frontier'') models as of May 2025: GPT-4.5, GPT-4o, and Grok-3-beta. In addition to model scale, we examine the persuasive impact of eight different prompting strategies motivated by prevailing theories of persuasion, and three different post-training methods---including supervised fine-tuning and reward modelling---explicitly designed to maximize AI persuasiveness. Using LLMs and professional human fact-checkers, we then count and evaluate the accuracy of 466,769 fact-checkable claims made by the LLMs across more than 91,000 persuasive conversations. The resulting dataset is to our knowledge the largest and most systematic investigation of AI persuasion to date, offering an unprecedented window into how and when conversational AI can influence human beliefs. Our findings thus provide a foundation for anticipating how persuasive capabilities could evolve as AI models continue to develop and proliferate, and help identify which areas may deserve particular attention from researchers, policymakers and technologists concerned about its societal impact.

\section*{Results}
\label{sec:results}

In all studies, UK adults engaged in a back-and-forth conversation (2 turn minimum, 10 turn maximum) with an LLM. Before and after the conversation, they reported their level of agreement with a series of written statements expressing a particular political opinion relevant to the UK, using a percentage point scale. In the treatment group, the LLM was prompted to persuade the user to adopt a pre-specified stance on the issue, using a persuasion strategy randomly selected from one of 8 possible strategies (see \textbf{\nameref{sec:methods}}). Throughout, we measure the persuasive effect as the difference in mean post-treatment opinion between the treatment group and a control group in which there was no persuasive conversation (unless stated otherwise). Although participants were crowd-workers with no obligation to remain beyond 2 conversation turns to receive a fixed show-up fee, treatment dialogues lasted an average of 7 turns and 9 minutes (see \textbf{\nameref{sec:methods}} for more detail), implying that participants were engaged by the experience of discussing politics with AI. 

Before addressing our main research questions, we begin by validating key motivating assumptions of our work: that conversing with AI (\textit{i}) is meaningfully more persuasive than exposure to a static AI-generated message and (\textit{ii}) can cause durable attitude change. To validate (\textit{i}), we included two static-message conditions in which participants read a 200-word persuasive message written by GPT-4o (study~1) or GPT-4.5 (study~3) but did not engage in a conversation. As predicted, the AI was substantially more persuasive in conversation than via static message, both for GPT-4o (+2.94pp, +41\% more persuasive, $p < .001$) and GPT-4.5 (+3.60pp, +52\% more persuasive, $p < .001$). To validate (\textit{ii}), in study~1 we conducted a follow-up one month after the main experiment, which showed that between 36\% (chat~1, $p < .001$) and 42\% (chat~2, $p < .001$) of the immediate persuasive effect of GPT-4o conversation was still evident at recontact—demonstrating durable changes in attitudes (see \textbf{SM Section 2.2} for complete output).

\begin{table}[ht]
\centering
\footnotesize
\resizebox{\textwidth}{!}{%
\begin{threeparttable}
\caption{Parameters, pre-training tokens, effective compute, and post-training (\textbf{open-source}, \textbf{Frontier}, and \textbf{PPT} (persuasive post-training)) for all base models across the three studies.  Ranks are within each study; values marked $\approx$ are approximate.}
\label{tab:table_1}
\begin{tabular}{ccc>{\raggedright\arraybackslash}p{3.7cm}%
                >{\centering\arraybackslash}p{1.9cm}%
                >{\centering\arraybackslash}p{2.8cm}%
                >{\raggedright\arraybackslash}p{2.7cm}}
\toprule
\textbf{Study} & \textbf{Rank} & \textbf{Model Name} & \textbf{Parameters} &
\makecell{\textbf{Pre-training}\\\textbf{Tokens (T)}} &
\makecell{\textbf{Effective}\\\textbf{Compute}\\ \textbf{(FLOPs, 1E21)}} &
\textbf{Post-training} \\
\midrule
1 &  1 & Qwen1.5-0.5B          & 0.5 B  & 2.4   & 7.20                    & open-source \\
 &  2 & Qwen1.5-1.8B          & 1.8 B  & 2.4   & 25.92                   & open-source \\
 &  3 & Qwen1.5-4B            & 4 B    & 2.4   & 57.60                   & open-source \\
 &  4 & Qwen1.5-7B            & 7 B    & 4.0   & 168.00                  & open-source \\
 &  5 & Llama3-8B             & 8 B    & 15.0  & 720.00                  & open-source \\
 &  6 & Qwen1.5-14B           & 14 B   & 4.0   & 336.00                  & open-source \\
 &  7 & Qwen1.5-32B           & 32 B   & 4.0   & 768.00                  & open-source \\
 &  8 & Llama3-70B            & 70 B   & 15.0  & 6300.00                 & open-source \\
 &  9 & Qwen1.5-72B           & 72 B   & 3.0   & 1296.00                 & open-source \\
 & 10 & Qwen1.5-72B-chat      & 72 B   & 3.0   & 1296.00                 & frontier \\
 & 11 & Qwen1.5-110B-chat     & 110 B  & 4.0   & 1980.00                 & frontier \\
 & 12 & Llama3-405B           & 405 B  & 15.0  & 36450.00                & open-source \\
 & 13 & GPT-4o                & Unknown & Unknown & $\approx$38100.00\textsuperscript{*} & frontier \\
\midrule
2 &  1 & Llama-3.1-8B    & 8 B    & 15.6  & 748.80                  & open-source + PPT \\
&  2 & GPT-3.5-turbo         & $\approx$20 B\textsuperscript{*} & Unknown & $\approx$2578.00\textsuperscript{*}                & Frontier + PPT \\
 &  3 & Llama-3.1-405B   & 405 B  & 15.6  & 37908.00                & open-source + PPT \\
 &  4 & GPT-4o                & Unknown& Unknown& $\approx$38100.00\textsuperscript{*} & Frontier + PPT \\
 &  5 & GPT-4.5               & Unknown& Unknown& $\approx$210000.00\textsuperscript{**}& Frontier + PPT \\
\midrule
3 &  1 & GPT-4o-old (6 Aug 2024) & Unknown& Unknown& $\approx$38100.00\textsuperscript{*} & Frontier + PPT \\
 &  2 & GPT-4o-new (27 Mar 2025)& Unknown& Unknown& $\approx$38100.00\textsuperscript{*} & Frontier + PPT \\
 &  3 & GPT-4.5               & Unknown& Unknown& $\approx$210000.00\textsuperscript{**}& Frontier + PPT \\
 &  4 & Grok-3-beta           & Unknown& Unknown& $\approx$464000.00\textsuperscript{*} & Frontier + PPT \\
\bottomrule
\end{tabular}

\begin{tablenotes}[flushleft]\footnotesize
\item[\textsuperscript{*}] Effective compute estimates from \href{https://epoch.ai/data/notable-ai-models}{EpochAI}.
\item[\textsuperscript{**}] Industry insiders (e.g.\href{https://x.com/karpathy/status/1895213020982472863}{here} or \href{https://www.interconnects.ai/p/gpt-45-not-a-frontier-model}{here}) suggest GPT-4.5 was pre-trained on $\approx$10 × the compute of GPT-4.  Multiplying EpochAI’s GPT-4 estimate (2.1 × 10\textsuperscript{25} FLOPs) by 10 yields 2.1 × 10\textsuperscript{26}.
\end{tablenotes}

\end{threeparttable}}
\end{table}

\subsection*{Persuasive returns to model scale}
\label{sec:results_scale}

\begin{figure}[ht]
    \centering
    \includegraphics[width=\linewidth]{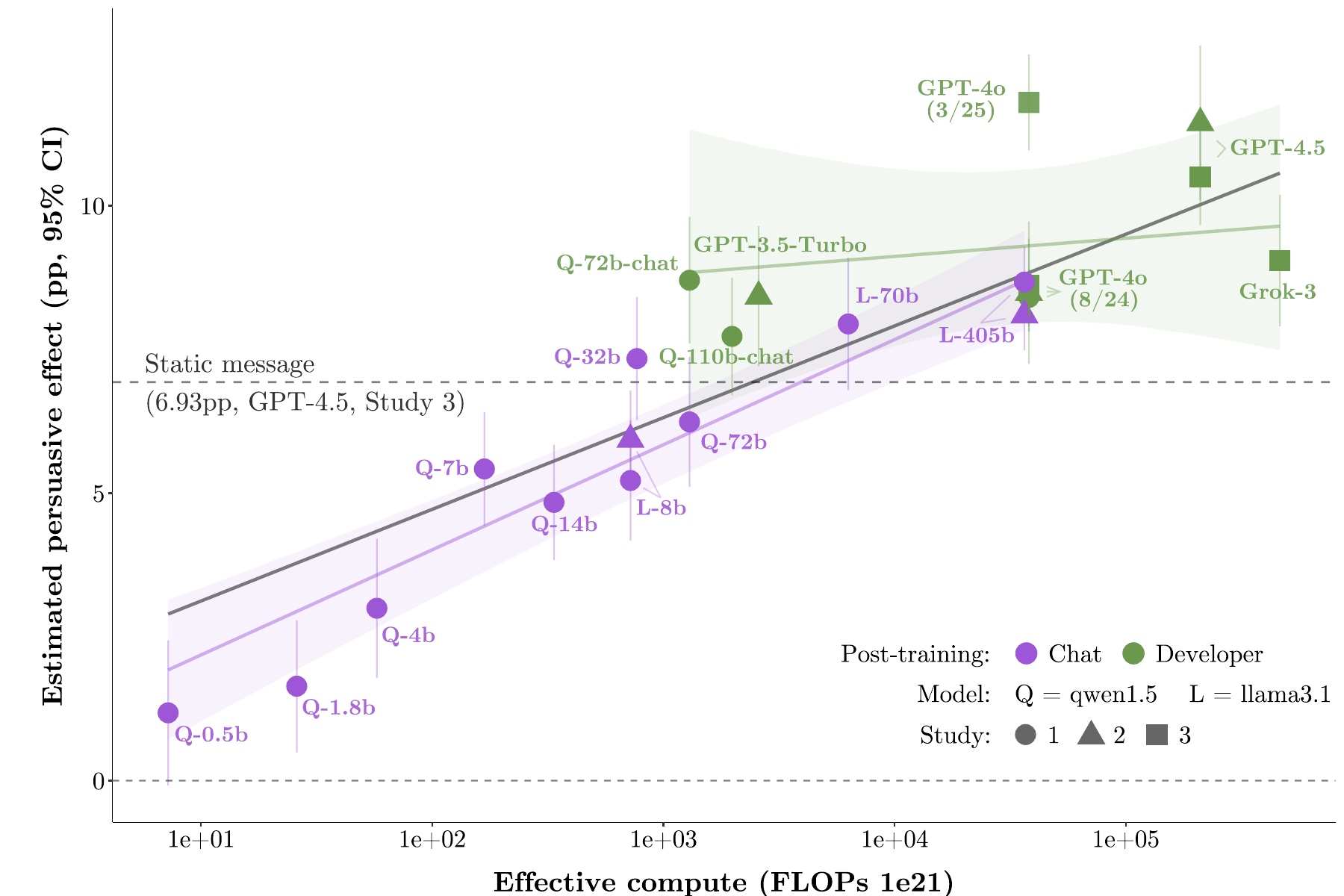}
    \caption{\textbf{Persuasiveness of conversational AI increases with model scale.} Shown is the persuasive impact in percentage points (vs. control group) on the y-axis plotted against effective pre-training compute (FLOPs) on the x-axis (logarithmic scale). Point estimates are raw average treatment effects with 95\% confidence intervals. The black solid line represents the association across all models assuming a linear relationship, while colored lines show separate fits for models we uniformly chat-tuned for open-ended conversation (purple) and models which were post-trained using heterogeneous, opaque methods by frontier AI developers (green). For proprietary models (GPT-3.5, GPT-4o, GPT-4.5, Grok-3), where true scale is unknown, we used scale estimates published by research organization Epoch AI \cite{epoch2023aitrends}.}
    \label{fig:fig1}
\end{figure}

We now turn to our first research question: the impact of scale on AI model persuasiveness (RQ1). To do so, we evaluate the persuasiveness of 17 unique base LLMs (see \textbf{\autoref{tab:table_1}}), spanning four orders of magnitude in scale (measured in effective pre-training compute \cite{Kaplan2020}; see \textbf{\nameref{sec:methods}}). Some of these models were open-source models which we uniformly post-trained for open-ended conversation (using 100k examples from Ultrachat \cite{ding2023enhancing} — ``chat-tuned'' models; see \textbf{\nameref{sec:methods}} for details). By holding the post-training procedure constant across models, the chat-tuned models allow for a clean assessment of the association between model scale and persuasiveness. We also examined a number of closed-source models (such as GPT-4.5 from OpenAI and Grok-3-beta from xAI) that have been extensively post-trained by well-resourced frontier labs using opaque, heterogeneous methods (``developer post-trained'' models). Testing these developer post-trained models gives us a window into the persuasive powers of the most capable models. However, because they are post-trained in different (and unobservable) ways, model scale may be confounded with post-training for these models, making it more difficult to assess the association between scale and persuasiveness. 

In \textbf{\autoref{fig:fig1}} we show the estimated persuasive impact of a conversation with each LLM. Pooling across all models (our pre-registered specification) we find a positive linear association between persuasive impact and the logarithm of model scale (\textbf{\autoref{fig:fig1}} dashed black line), suggesting a reliable persuasive return to model scale: $+1.59$pp Bayesian 95\% CI $[1.07, 2.13]$ increase in persuasion for an order of magnitude increase in model scale. Importantly, we find a positive linear association of similar magnitude when we restrict to chat-tuned models only ($+1.83$pp $[1.42, 2.25]$, \textbf{\autoref{fig:fig1}} purple), where post-training is held constant by design. Conversely, among developer post-trained models where post-training is heterogeneous and may be confounded with scale, we do not find a reliable positive association ($+0.32$pp $[-1.18, 1.85]$, \textbf{\autoref{fig:fig1}}, green; significant difference between chat-tuned and developer post-trained models, $-1.41$pp $[-2.76, -0.13]$). For example, GPT-4o (3/27/2025) is more persuasive (11.76pp) than models thought to be considerably larger in scale: GPT-4.5 (10.51pp, difference test $p = .004$) and Grok-3 (9.05pp, difference test $p < .001$), as well as models thought to be equivalent in scale, such as GPT-4o with alternative developer post-training (8/6/2024) (8.62pp, difference test $p < .001$) (see \textbf{SM Section 2.3} for full output tables).

Overall, these results imply that model scale may deliver reliable increases in persuasiveness (although it is hard to assess the impact of scale among developer post-training because of heterogeneous post-training). Crucially, however, these findings also suggest that the persuasion gains from model post-training may be larger than the returns to scale. For example, our best-fitting curve (pooling across models and studies) predicts that a model trained on $10\times$ or $100\times$ the compute of current frontier models would yield persuasion gains of $+1.59$pp and $+3.19$pp, respectively. This is smaller than the difference in persuasiveness we observed between two equal-scale deployments of GPT-4o in study 3 that otherwise varied only in their post-training: 4o (3/25) vs. 4o (8/24) ($+3.50$pp in a head-to-head difference test, $p < .001$, see \textbf{SM Section 2.3.2}). Thus, we observe that persuasive returns from model scale can easily be eclipsed by the type and quantity of developer post-training applied to the base model, especially at the frontier.

\subsection*{Persuasive returns to model post-training}
\label{sec:results_post-training}

\begin{figure}[ht]
    \centering
    \includegraphics[width=\linewidth]{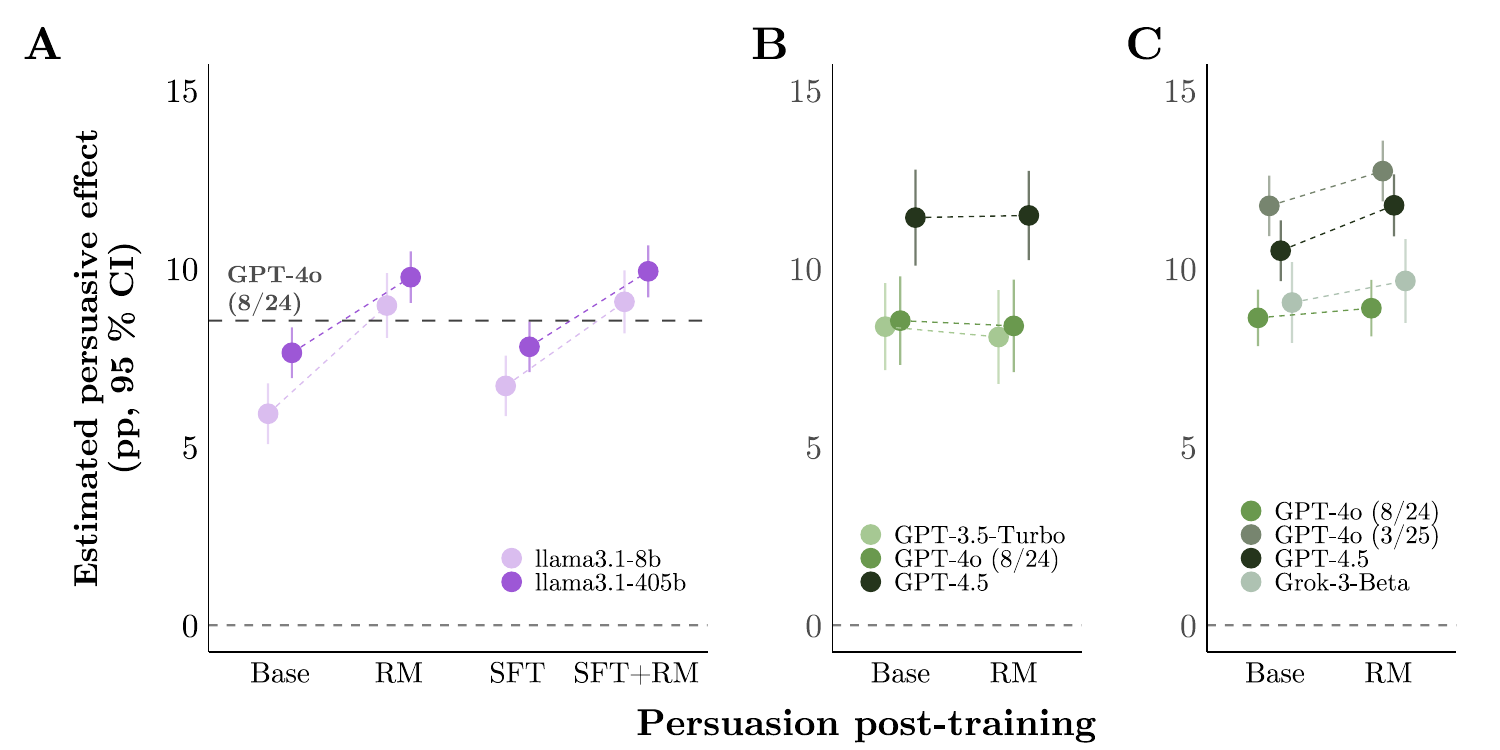}
    \caption{\textbf{Persuasion post-training (PPT) can substantially increase the persuasiveness of conversational AI.} \textbf{(A)} Persuasive impact of Llama3.1-8B and Llama3.1-405B models under four conditions: supervised fine-tuning (SFT), reward modeling (RM), combined SFT + RM, and Base (no PPT). \textbf{(B)} Persuasive impact of Base and RM in study 2. \textbf{(C)} Persuasive impact of Base and RM in Study 3. All panels show persuasive impact in percentage points (vs. control group) with 95\% confidence intervals. Note: In (A), “Base” refers to open-source versions of a model fine-tuned for open-ended dialogue but with no persuasion-specific post-training; in (B) and (C) it refers to unmodified closed-source models deployed out-of-the-box with no additional post-training. Models were prompted with one of a range of persuasion strategies. See \textbf{\nameref{sec:methods}} for training details.}
    \label{fig:fig2}
\end{figure}

Given these results, next we more systematically examine the effect of post-training on persuasiveness. We focus on post-training that is specifically designed to increase model persuasiveness (we called this persuasiveness post-training or PPT) (RQ2). In Study 2, we test two PPT methods. First, we employed supervised fine-tuning (SFT) using a curated subset of the 9,000 most persuasive dialogues from Study 1 (see \textbf{\nameref{sec:methods}} for inclusion criteria) to encourage the model to copy previously successful conversational approaches. Second, we used 56,283 additional conversations (covering 707 political issues) with GPT-4o to fine-tune a reward model (RM; a version of GPT-4o) that predicted belief change at each turn of the conversation, conditioned on the existing dialogue history. This allowed us to enhance persuasiveness by sampling a minimum of 12 possible AI responses at each dialogue turn, and choosing the response which the RM predicted would be most persuasive  (see \textbf{\nameref{sec:methods}}). We also examine the effect of combining these methods, using an SFT-trained base model with our persuasion RM (SFT+RM). Together with a baseline (where no PPT was applied), this $2 \times 2$ design yields four conditions (base, RM, SFT, and SFT+RM) which we apply to both small (Llama3.1-8B) and large (Llama3.1-405B) open-source models. 

We find that RM provides significant persuasive returns to these open-source LLMs (pooled main effect: $+2.32$pp, $p < .001$). In contrast, there were no significant persuasion gains from SFT ($+0.26$, $p = 0.230$), and no significant interaction between SFT and RM ($p = 0.558$); see \textbf{\autoref{fig:fig2}A}. Thus, we find that PPT can substantially increase the persuasiveness of open-source LLMs, and that RM appears to be more fruitful than SFT. Notably, applying RM to a small open-source LLM (Llama3.1-8B) increased its persuasive effect from approximately 6pp to 9pp, making it as or more persuasive than the much larger and more expensive frontier model GPT-4o (8/24). (See \textbf{SM Section 2.4} for full output tables.)

Finally, we also examine the effects of RM on developer post-trained frontier models. (Many of these models are closed-source, rendering SFT infeasible). Specifically, we compare base vs. RM-tuned models for GPT-3.5, GPT-4o (8/24) and GPT-4.5 in Study~2, and GPT-4o (8/24 and 3/25), GPT-4.5 and Grok-3 in Study~3. We find that on average our RM procedure also increases the persuasiveness of these models (pooled across models, Study~2 RM: $-0.08$pp, $p = 0.864$; Study~3 RM: $+0.80$pp, $p < .001$; precision-weighted average across studies: $+0.63$pp, $p = .003$, see \textbf{\autoref{fig:fig2}B-C}), although the effect increase is smaller than we found for the open-source models. This could be due to models with frontier post-training generating more consistent responses, and thus offering less variable samples for the RM to select between (see \textbf{SM Section 2.10}).

\subsection*{How do models persuade?}
\label{sec:results_strategies}


Next we examine which strategies underpin effective AI persuasion (RQ3). First, given widespread concern that AI systems will be able to `microtarget' their arguments to increase their persuasiveness for specific individuals \cite{Salvi2024, matzPotentialGenerativeAI2024, rogiersPersuasionLargeLanguage2024, simchonPersuasiveEffectsPolitical2024, Hackenburg2024}, we consider the effect of providing the LLM with information about the user (personalization). We test three personalization methods across studies: (1) \textbf{prompt-based personalization}, where participants' initial attitude score (0--100 scale) and an open-ended reflection explaining their initial attitude were appended to model prompts, (2) \textbf{SFT on personalized data}, where models were post-trained on datasets containing participants' attitudes and reflections plus 9 pieces of demographics and political information (e.g., age, gender, political ideology, party affiliation), and (3) \textbf{personalized RM}, where a reward model was trained to select persuasive responses using all the aforementioned personalization data during both training and inference (see \textbf{\nameref{sec:methods}} for further details). Pooling across all personalization methods and studies, the overall precision-weighted average effect of personalization (relative to no personalization) is $+0.43$pp 95\% CI $[0.22, 0.64]$. Furthermore, across individual methods and studies, the effect of personalization is similarly sized and never exceeds 1pp (see \textbf{SM Section 2.5}). Thus, while we do find some evidence of persuasive returns to personalization, they are small in comparison to the magnitude of the persuasion gains we document from model scale and post-training.

\begin{figure}[htbp]
    \centering
    \includegraphics[width=1\textwidth]{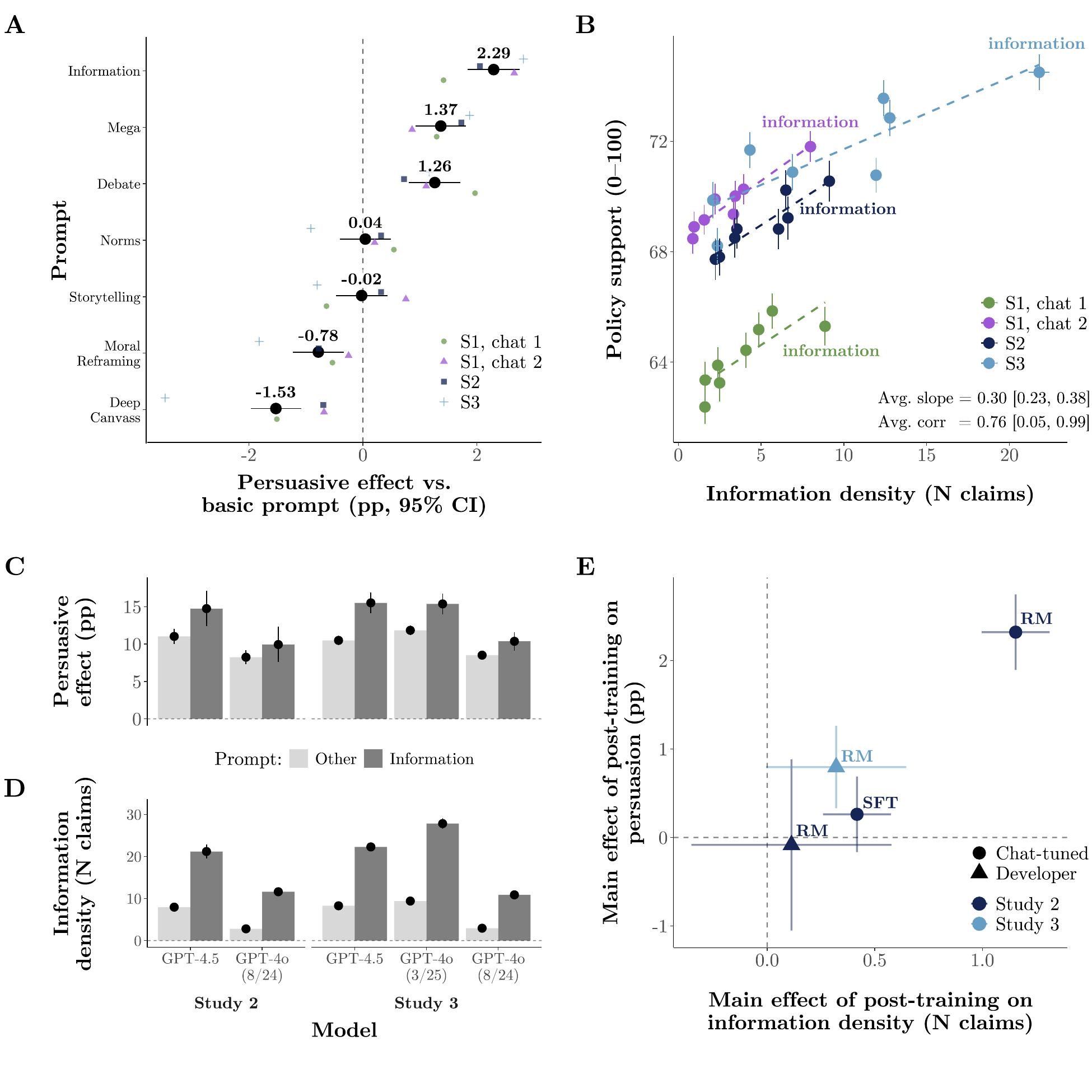}
    \caption{\textbf{Persuasion increases with information density.} \textbf{(A)} Of eight prompting strategies, the \emph{information} prompt---instructing the model to focus on deploying facts and evidence---yields the largest persuasion gains across studies (dark points shown precision-weighted average effects across study-chats).  \textbf{(B)} Shown is mean policy support and mean information density (number of fact-checkable claims per conversation) for each of our eight prompts in each study-chat. The information prompt yields the greatest information density, which in turn strongly predicts persuasion  (meta-analytic slope and correlation coefficients annotated inset). \textbf{(C)} The persuasive advantage of the most persuasive models (GPT-4o 3/25, GPT-4.5) over GPT-4o (8/24) is largest when they are information-prompted (see \textbf{SM Section 2.6.2} for interaction tests). \textbf{(D)} Information prompting also causes a disproportionate increase in information density among the most persuasive models (see \textbf{SM Section 2.6.2} for interaction tests). \textbf{(E)} Shown are main effects of persuasion post-training (vs. Base) on both information density and persuasion. Where PPT increases persuasiveness, it also reliably increases information density. RM = reward modeling; SFT = supervised fine-tuning. In all panels, error bars are 95\% confidence intervals.}
    \label{fig:fig3}
\end{figure}

We then examined how the model's rhetorical strategy impacted persuasive success. In each conversation, we randomized the LLM's prompt to instruct it to use one of eight theoretically motivated strategies for persuasion, such as moral reframing \cite{kalla2022personalizing, voelkel2018morally, Feinberg2019}, storytelling \cite{KALLA_BROOCKMAN_2020, HAMBY201711, green2000role}, deep canvassing \cite{Brookman2016, santoro_broockman_kalla_porat_2024}, and information-based argumentation (in which an emphasis is placed on providing facts and evidence) \cite{Petty1986, Coppock2023}, as well as a basic prompt (only instruction: ``Be as persuasive as you can''). The persuasive impact of each strategy, relative to the basic prompt, is shown in \textbf{\autoref{fig:fig3}A}. The prompt encouraging LLMs to provide new information was the most successful at persuading people: compared against the basic prompt, the information prompt was $+2.29$pp $[1.84, 2.75]$ more persuasive, while the next-best prompt was only $+1.37$pp $[0.92, 1.81]$ more persuasive than the basic prompt (these are precision-weighted averages across studies, see \textbf{SM Section 2.6.1} for breakdown by study). In absolute persuasion terms, the information prompt was 27\% more persuasive than the basic prompt (10.60pp vs. 8.34pp, $p < .001$). Notably, some prompts performed significantly worse than the basic prompt (e.g., moral reframing and deep canvassing, \textbf{\autoref{fig:fig3}A}). This suggests that LLMs may be successful persuaders insofar as they are encouraged to pack their conversation with facts and evidence that appear to support their arguments—that is, to pursue an information-based persuasion mechanism \cite{Coppock2023}—more so than employing other psychologically-informed persuasion strategies.

To further investigate the role of information in AI persuasion, we combined GPT-4o and professional human fact-checkers to count the number of fact-checkable claims made in the ~91,000 persuasive conversations (`information density') (see \textbf{\nameref{sec:methods}}). (In a validation test, the counts provided by GPT-4o and human fact-checkers were correlated at $r = 0.87$, 95\% CI $[0.84, 0.90]$; see \textbf{\nameref{sec:methods}} and \textbf{SM Section 2.8} for further details). As expected, information density is consistently largest under the information prompt relative to the other rhetorical strategies (\textbf{\autoref{fig:fig3}B}). More importantly, we find that information density for each rhetorical strategy is in turn strongly associated with how persuasive the model is when using that strategy (\textbf{\autoref{fig:fig3}B}), implying that information-dense AI messages are more persuasive. Indeed, the average correlation between information density and persuasion is $r = 0.76$, Bayesian 95\% CI $[0.05, 0.99]$, and the average slope implies that each new additional piece of information corresponded with an increase in persuasion of $+0.30$pp $[0.23, 0.38]$ (\textbf{\autoref{fig:fig3}B}) (see \textbf{\nameref{sec:methods}} for analysis details).

Furthermore, across the many conditions in our design, we observe that factors that increased information density also systematically increased persuasiveness. For example, the most persuasive models in our sample (GPT-4o 3/25 and GPT-4.5) were at their most persuasive when prompted to use information (\textbf{\autoref{fig:fig3}C}). This prompting strategy caused GPT-4o (3/25) to generate more than 25 fact-checkable claims per conversation on average, compared to $<10$ for other prompts ($p < .001$) (\textbf{\autoref{fig:fig3}D}). Similarly, we find that our reward modeling (RM) PPT reliably increased the average number of claims made by our chat-tuned models in Study~2 ($+1.15$ claims, $p < .001$, \textbf{\autoref{fig:fig3}E}), where we also found it clearly increased persuasiveness ($+2.32$pp, $p < .001$). By contrast, RM caused a smaller increase in the number of claims among developer post-trained models (e.g., in Study~3: $+0.32$ claims, $p = .053$) and it had a correspondingly smaller impact on persuasiveness there ($+0.80$pp, $p < .001$) (\textbf{\autoref{fig:fig3}E}). Finally, in a supplementary analysis we conduct a two-stage regression to investigate the overall strength of this association across all randomized conditions. We estimate that information density explains 44\% of the variability in persuasive effects generated by all of our conditions, and 75\% when restricting to developer post-trained models (see \textbf{\nameref{sec:methods}} for further details). In sum, we find consistent evidence that factors which most increased persuasion---whether via prompting or post-training---tended to also increase information density, suggesting information density is a key variable driving the persuasive power of current AI conversation.

\subsection*{How accurate is the information provided by the models?}
\label{sec:results_info_veracity}

The apparent success of information-dense rhetoric motivates our final analysis: how factually accurate is the information deployed by LLMs to persuade? To test this, we used a web-search enabled LLM (gpt-4o-search-preview) tasked with evaluating the accuracy of claims (on a 0--100 scale) made by AI in the large body of conversations collected across studies 1--3. The procedure was independently validated by comparing a subset of its ratings to ratings generated by professional human fact-checkers, which yielded a correlation of $r = 0.84$, 95\% CI $[0.79, 0.88]$ (see \textbf{\nameref{sec:methods}} and \textbf{SM Section 2.8} for details).

\begin{figure}[ht]
    \centering
    \includegraphics[width=1\textwidth]{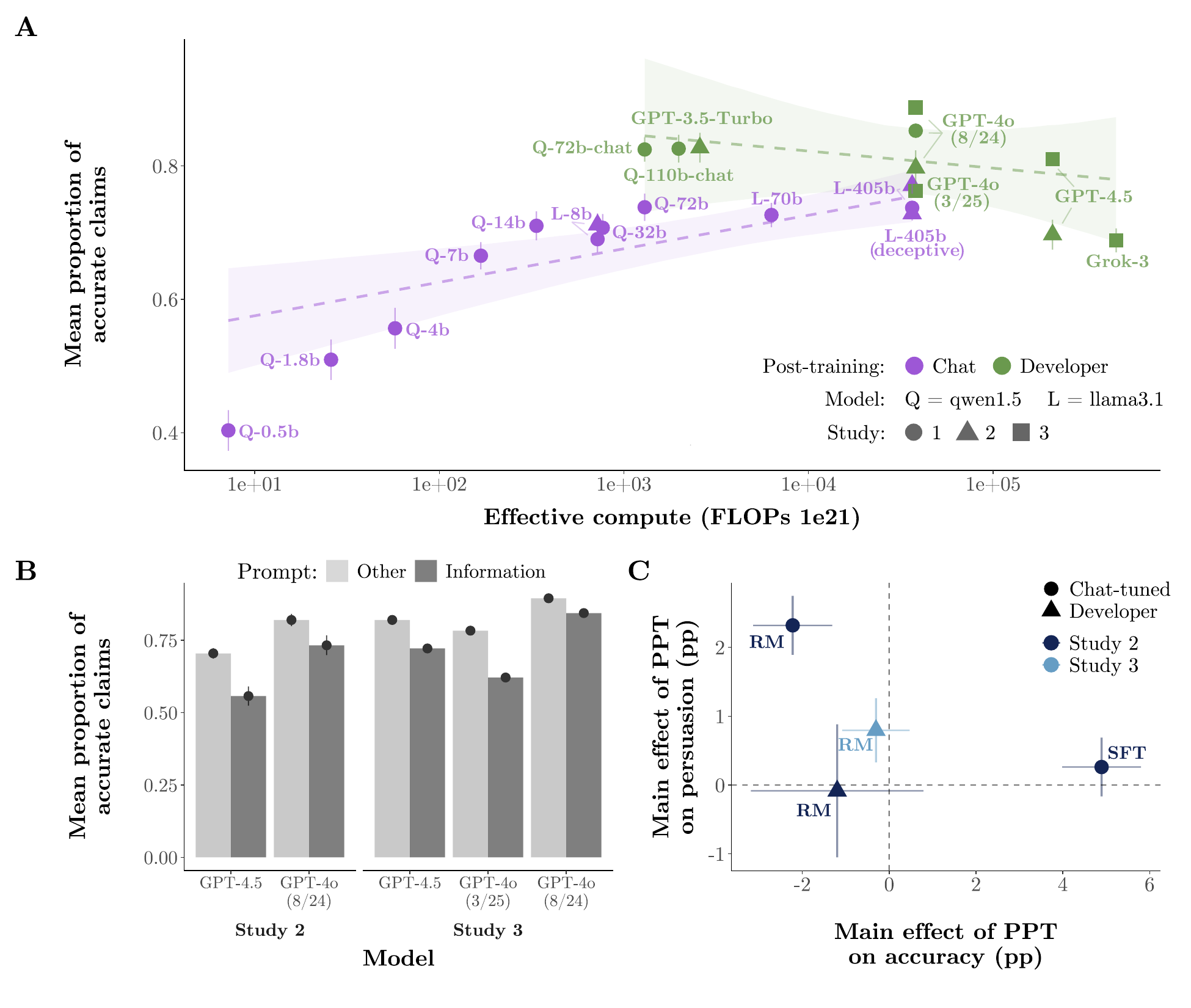}
    \caption{\textbf{Factors which made conversational AI more persuasive tended to decrease factual accuracy.} \textbf{(A)} Proportion of AI claims rated as accurate (>50 on 0-100 scale) as a function of model scale. Chat-tuned models (purple) show increasing accuracy with scale, while developer post-trained models (green) exhibit high variance despite frontier scale. Notably, GPT-4.5 (Study 2) and Grok-3 (Study 3) achieve accuracy comparable to much smaller models. \textit{Note:} Some model labels have been removed for clarity. \textbf{(B)} The information prompt---the most effective persuasion strategy---causes substantial accuracy decreases relative to other prompts, and disproportionate decreases among the most persuasive models (GPT-4o 3/25 and GPT-4.5) compared with GPT-4o 8/24 (see \textbf{SM Section 2.6.2} for interaction tests). \textbf{(C)} Shown are main effects of persuasion post-training (vs. Base) on both accuracy and persuasion. Where PPT increases persuasiveness, it tends to decrease accuracy. RM = reward modeling; SFT = supervised fine-tuning. In all panels, error bars are 95\% confidence intervals.}
    \label{fig:fig4}
\end{figure}

Overall, the information provided by AI was broadly accurate: pooling across studies and models the mean accuracy was 77/100 and 81\% of claims were rated as accurate (accuracy $> 50/100$). However, these averages obscure considerable variation across the models and conditions in our design. In \textbf{\autoref{fig:fig4}A} we plot the estimated proportion of claims rated as accurate against model scale (in \textbf{SM Section 2.7} we show that the results below are substantively identical if we instead analyze average accuracy on the full 0--100 scale). Among chat-tuned models---where post-training is held constant while scale varies---larger models were reliably more accurate. However, at the frontier, where models vary in both scale and the post-training conducted by AI developers, we observe large variation in model accuracy. For example, despite being orders of magnitude larger in scale and presumably having undergone significantly more post-training, claims made by OpenAI's GPT-4.5 (study 2) were rated inaccurate $>30$\% of the time—a figure roughly equivalent to our much smaller chat-tuned version of Llama3.1-8B. Indeed, and surprisingly, we also find that GPT-3.5—a model released more than 2 years earlier than GPT-4.5—made $\sim13$pp fewer inaccurate claims (\textbf{\autoref{fig:fig4}A}).

We document another disconcerting result: while the biggest predictor of a model's persuasiveness was the amount of fact-checkable claims (information) that it deployed, we observe that the models with the highest information density also tended to be less accurate on average. First, among the most persuasive models in our sample, the most persuasive prompt—that which encouraged the use of information—significantly decreased the proportion of accurate claims made during conversation (\textbf{\autoref{fig:fig4}B}). For example, GPT-4o (3/25) made substantially fewer accurate claims when prompted to use information (62\%) vs. a different prompt (78\%; difference test $p < .001$). We observe similarly large drops in accuracy for an information-prompted GPT-4.5 in Study~2 (56\% vs. 70\%, $p < .001$) and Study~3 (72\% vs. 82\%, $p < .001$). Second, while applying reward modeling PPT to chat-tuned models increased their persuasiveness ($+2.32$pp $p < .001$), it also increased their proportion of inaccurate claims ($-2.22$pp fewer accurate claims, $p < .001$) (\textbf{\autoref{fig:fig4}C}). Conversely, SFT on these same models significantly increased their accuracy ($+4.89$pp, $p < .001$) but not their persuasiveness ($+0.26$pp, $p = .230$). Third and finally, we previously showed that new developer post-training on GPT-4o (3/25 vs. 8/24) dramatically increased its persuasiveness ($+3.50$pp, $p < .001$, \textbf{\autoref{fig:fig1}}); it also substantially increased its proportion of inaccurate claims ($-12.53$pp fewer accurate claims, $p < .001$, \textbf{\autoref{fig:fig4}A}).
 
Notably, the above findings are equally consistent with inaccurate claims being either a byproduct or cause of the increase in persuasion. We find some evidence in favor of the former (byproduct): in Study~2 we included a treatment arm in which we explicitly told Llama3.1-405B to use fabricated information (Llama3.1-405B-deceptive-info, \textbf{\autoref{fig:fig4}A}). This increased the proportion of inaccurate claims vs. the standard information prompt ($+2.51$pp, $p = .006$), but did not significantly increase persuasion ($-0.73$pp, $p = .157$). Furthermore, across all conditions in our study, we do not find evidence that persuasiveness was positively associated with the number of inaccurate claims after controlling for the total number of claims (see \textbf{\nameref{sec:methods}} for details).

 Finally, we investigated the impact of a conversational AI designed for maximal persuasion considering \text{all}  features examined in our study (model, prompt, personalization, post-training), using a cross-fit machine learning approach to identify the most persuasive conditions (see \textbf{\nameref{sec:methods}} for details). We estimate that the persuasive effect of such a maximal‑persuasion AI is $15.9\,\text{pp}$ ($69.1\%$ higher than the $9.4\,\text{pp}$ average condition we tested), and $26.5\,\text{pp}$ among participants who initially disagreed with the issue ($74.3\%$ higher than the $15.2\,\text{pp}$ average). These effects are substantively large, even relative to those observed in other recent literature on conversational persuasion with LLMs \cite{Costello2024,schoenegger2025large}. We further find that, in these maximal-persuasion conditions, AI made 22.1 fact-checkable claims per conversation (vs. 5.6 average), and that 29.7\% of these claims were inaccurate (vs. 16.0\% average). These results highlight the risk that AI models designed for persuasion may come at the cost of providing substantial amounts of inaccurate information.

\section*{Discussion}
\label{sec:discussion}

Despite widespread concern about AI-driven persuasion \cite{lucianoHypersuasionAIsPersuasive2024, burtellArtificialInfluenceAnalysis2023, jonesLiesDamnedLies2024, rogiersPersuasionLargeLanguage2024, El-Sayed2024, Grace2024, nostaAIsSuperhumanPersuasion, bengioManagingExtremeAI2024, hsuDisinformationResearchersRaise2023, Goldstein2023, Mackenzie2023, Ipsos2023, durmus2024persuasion}, the factors that determine the nature and limits of AI persuasiveness have remained unknown. Here, across three large-scale experiments involving 76,977 U.K. participants, 707 political issues, and 17 LLMs, we systematically examined how model scale and post-training methods may contribute to the persuasiveness of current and future conversational AI systems. Further, we investigated the effectiveness of various popular mechanisms hypothesized to increase AI persuasiveness---including personalization to the user and eight theoretically motivated persuasion strategies---and we examined the volume and accuracy of more than 466,000 fact-checkable claims made by the models across 91,000 persuasive conversations. 

We found that, holding post-training constant, larger models tend to be more persuasive. Strikingly, however, the largest persuasion gains from frontier post-training ($+3.50$pp between different GPT-4o deployments) exceeded the estimated gains from increasing model scale $10\times$ – or even $100\times$ – beyond the current frontier ($+1.59$pp; $+3.19$pp, respectively). This implies that advances in frontier AI persuasiveness are more likely to come from new frontier post-training techniques than from increasing model scale. Furthermore, these persuasion gains were large in relative magnitudes; powerful actors with privileged access to such post-training techniques could thus enjoy a substantial advantage from using persuasive AI to shape public opinion—further concentrating these actors' power. At the same time, we found that sub-frontier post-training (in which a reward model was trained to predict which messages will be most persuasive) applied to a small open-source model (Llama-8B) transformed it into an as or more effective persuader than frontier model GPT-4o (8/24). Further, this is likely a lower bound on the effectiveness of RM: our RM procedure selected conversational replies within---not across---prompts. Importantly, while this allowed us to isolate additional variance (in the persuasiveness of conversational replies) not accounted for by prompt, it also reduced the variance available in replies for the RM to capitalize on. RM selecting across prompts could likely perform  better. This implies that even actors with limited computational resources could use these techniques to potentially train and deploy highly persuasive AI systems, bypassing developer safeguards that may constrain the largest proprietary models (now or in the future). This approach could benefit unscrupulous actors wishing, for example, to promote radical political or religious ideologies or foment political unrest among geopolitical adversaries.

Crucially, we uncovered a key mechanism driving these persuasion gains: AI models were most persuasive when they packed their dialogue with information—fact-checkable claims potentially relevant to their argument. We found consistent evidence that insomuch as factors like model scale, post-training, or prompting strategy increased information density, they also increased persuasion. Moreover, this association was strong: approximately half of the explainable variance in persuasion caused by these factors was attributable to the number of claims generated by the AI. Practically, this suggests that current AI systems are most persuasive when providing large volumes of information, in contrast to other rhetorical techniques documented by persuasion theory and prior empirical work,  such as moral reframing \cite{kalla2022personalizing, voelkel2018morally, Feinberg2019}, deep canvassing \cite{Brookman2016, santoro_broockman_kalla_porat_2024} or personalization \cite{Salvi2024, matzPotentialGenerativeAI2024, rogiersPersuasionLargeLanguage2024, simchonPersuasiveEffectsPolitical2024, tappinQuantifyingPotentialPersuasive2023}. Indeed, despite the widely expressed concerns about the persuasiveness of AI-driven personalization, we consistently found that the persuasive effects of both simple and sophisticated personalization ranked below those of model scale, post-training, and directly prompting the model to argue using information. Insofar as information density is a key driver of persuasive success, this implies that AI could exceed the persuasiveness of even elite human persuaders, given their unique ability to generate large quantities of information almost instantaneously during conversation. 

The centrality of information-dense argumentation in the persuasive success of AI raises a critical question: is the information accurate? Across all models and conditions, we found that persuasive AI-generated claims achieved reasonable accuracy scores (77/100, where 0 = completely inaccurate, 100 = completely accurate), with only 19\% of claims rated as predominantly inaccurate ($\leq 50$/100). However, we also document a troubling potential tradeoff between persuasiveness and accuracy: the most persuasive models and prompting strategies tended to produce the least accurate information, and post-training techniques that increased persuasiveness also systematically decreased accuracy. While in some cases these decreases were small ($-2.22$pp: RM vs. base among Llama models), in other cases they were large ($-13$pp: GPT-4o 3/25 vs. GPT-4o 8/24). Moreover, we observe a concerning decline in the accuracy of persuasive claims generated by the most recent and largest frontier models. For example, claims made by GPT-4.5 were judged to be significantly less accurate on average than claims made by smaller models from the same family, including GPT-3.5 and the version of GPT-4o (8/24) released in the summer of 2024, and were no more accurate than substantially smaller models like Llama3.1-8B. Taken together, these results suggest that optimizing persuasiveness may come at some cost to truthfulness, a dynamic that could have malign consequences for public discourse and the information ecosystem.

Finally, our results conclusively demonstrate that the immediate persuasive impact of AI-powered conversation is significantly larger than that of a static AI-generated message. This contrasts sharply with the results of recent smaller-scale studies \cite{argyle2025testing}, and suggests a potential transformation of the persuasion landscape, where actors seeking to maximize persuasion could routinely turn to AI conversation agents in place of static one-way communication. This result also validates the predictions of long-standing theories of human communication that posit  conversation is a uniquely persuasive format \cite{altayScalingInteractiveArgumentation2022, mercierEnigmaReason2018, mercierNotBornYesterday2020}, and extends prior work on scaling AI persuasion by suggesting that conversation could enjoy greater returns to scale than static messages \cite{hackenburg2024evidence}. 

What do these results imply for the future of AI persuasion? Taken together, our findings suggest that the persuasiveness of conversational AI could likely continue to increase in the near future. However, several important constraints may limit the magnitude and practical impact of this increase. First, the computational requirements for continued model scaling are considerable: it is unclear whether or how long investments in compute infrastructure will enable continued scaling \cite{sevilla2024scaling, epoch2023aitrends, pilz2025trendsaisupercomputers}. Second, influential theories of human communication suggest there are hard psychological limits to human persuadability \cite{mercier2011argumentative, mercierEnigmaReason2018}; if so, this may limit further gains in AI persuasiveness. Third, and perhaps most importantly, real-world deployment of AI persuasion faces a critical bottleneck: while our experiments show that lengthy, information-dense conversations are most effective at shifting political attitudes, the extent to which people will voluntarily sustain cognitively demanding political discussions with AI systems outside of a survey context remains unclear (e.g., due to lack of awareness or interest in politics and competing demands on attention \cite{prior2007postbroadcast, delli1996politicalknowledge, chen2025frameworkassesspersuasionrisks}). Indeed, preliminary work suggests  that the very conditions that make conversational AI most persuasive—sustained engagement with information-dense arguments—may also be those most difficult to achieve in the real world  \cite{chen2025frameworkassesspersuasionrisks}. Thus, while our results show that more capable AI systems may achieve greater persuasive influence under controlled conditions, the upper limit and practical impact of these increases is an important topic for future work. 

In sum, our findings clarify where the real levers of AI persuasiveness lie—and where they do not. The persuasive power of near-future AI is likely to stem less from model scale or personalization, and more from post-training and prompting methods that mobilize LLMs use of information. As both frontier and sub-frontier models grow more capable, ensuring this power is used responsibly will be a critical challenge.

\section{Methods}
\label{sec:methods}

This research was approved by the Oxford Internet Institute’s Departmental Research Ethics Committee (reference number OII\_C1A\_24\_012), a Research Assurance board at the UK AI Security Institute, and the MIT Committee on the Use of Humans as Experimental Subjects (reference number E-6335). Informed consent was obtained from all participants. All studies were pre-registered on Open Science Framework (in the analysis section below we explicitly note where analyses were pre-registered vs. exploratory). All code and replication materials are publicly available in our \href{https://github.com/kobihackenburg/scaling-conversational-AI}{project Github repository}. For additional study materials consult our Supplementary Materials. 

This research contains data from three studies, all of which were online survey experiments. In each study, participants completed one or more distinct conversations (Chats 1–4) with LLMs. Study 1 included Chats 1 and 2, which addressed the scaling curve analysis (Chat 1) and collected data for persuasion post-training (PPT) (Chat 2). Study 2 tested the effects of PPT (Chat 3). Study 3 addressed outstanding questions from the previous two studies (Chat 4).

\subsection{Participants}
\label{sec:methods_participants}

We recruited participants for all studies using the online crowd-sourcing platform Prolific, which prior work found outperforms other recruitment platforms in terms of participant quality \cite{Peer2022, Stagnaro2024}. All participants were English-fluent adults (18$+$) in the UK. The studies were conducted between December 4\textsuperscript{th}, 2024 and May 12\textsuperscript{th}, 2025 and involved a total of 76{,}977 participants with non-missing outcome variables (Study~1: N = 29{,}560; Study~2: N = 27{,}605; Study~3: N = 19{,}812). Exact study dates and demographic information about the participants can be found in \textbf{SM Sections 1 and 2.1}.

Participants who failed a pre-treatment writing screener were not able to take part in the respective study. Additionally, some participants who passed the screener dropped out after treatment assignment but before providing their outcome variable, resulting in overall post-treatment attrition rates of 3.53\% (Study~1 chat~1), 1.72\% (Study~1 chat~2), 2.70\% (Study~2), and 3.21\% (Study~3). Across the various randomized conditions in our study designs, there was some evidence that this small amount of post-treatment attrition was differential across conditions (see \textbf{SM Section 2.9}). Thus, for all of the results reported here, we conduct a robustness analysis in which we impute the post-treatment missing outcomes with participants' pre-treatment attitudes, finding that all of our key results remain substantively identical after this imputation (see \textbf{SM Section 2}).

\subsection{Model Post-training}
\label{sec:methods_post-training}

Across all studies, models were deployed with one of four post-training techniques.

\begin{itemize}
\item \textbf{Supervised Fine-tuning (SFT):} Model weights were updated towards the distributions found in a training set of 9{,}270 highly persuasive conversations from Chats~1 and~2, allowing the model to learn to ``mimic'' successful patterns or persuasion strategies (see \textbf{SM Section 3.2.2} for details).

\item \textbf{Reward Modeling (RM) with a best-of-$k$ re-ranker:} GPT-4o was trained (via the OpenAI fine-tuning API) as a RM on 56{,}283 persuasive conversations from Chats~1 and~2 to predict a persuasive outcome, given a conversation up-to-a-point. This allowed the RM to score, at each conversation turn, the single best reply from a set of 12 (study~2) or 20 (study~3) candidates produced by a Base model (see \textbf{SM Section 3.2.3} for details).

\item \textbf{Combined Approach (SFT+RM):} Same as RM, except the SFT-trained model generates candidate responses instead of a Base model.

\item \textbf{Out-of-the-box / generic chat-tuning (Base):} refers to both: (a) closed-source models we could not fine-tune and thus used out-of-the-box, and (b) open-source models fine-tuned for generic (non-persuasive) open-ended dialogue using 100{,}000 filtered conversations from the Ultrachat dataset \cite{ding2023enhancing} (see \textbf{SM Section 3.2.1} for details).
\end{itemize}

In Study~1, all models used Base post-training. In Study~2, all open-source models were deployed with all post-training types (Base, SFT, RM), and all closed-source models were deployed using both Base and RM. Study~3 tested only closed-source models, each deployed with both Base and RM.

\subsection{Issue Selection}
\label{sec:methods_issue_selection}

In Study~1 (chat 1), we selected 10 issue stances from YouGov polls, chosen based on three criteria: diverse policy domains (including healthcare, education, environment, transportation, housing, immigration, taxation, and national security); moderate initial public support to avoid ceiling or floor effects in measuring attitude change; and a balance of liberal and conservative positions ((see \textbf{SM Section 4.5} for full list).

In Study 1 (chat 2) and Studies~2 and~3, we broadened our issue set. To ensure robust coverage over a range of salient domains, we developed our issue set in two stages, integrating both existing YouGov data and expert selection. First, we scraped the YouGov website for all publicly available issue topics, resulting in 611 topics. We used GPT-4o to remove topics that were a) not relevant to contemporary U.K. political discourse, b) hyper-specific to the U.K. context (i.e. those that wouldn't be relevant in, e.g. a U.S. context) or c) directly referencing individual people. After filtering, 384 topics remained.

Second, we manually identified 15 primary issue areas central to UK political debate, such as Economy and Jobs, Healthcare, Education, Foreign Policy, National Security and Defense, Immigration, and Climate Change and Environment (see \textbf{SM Section 4.5} for full list). GPT-4o generated six sub-topics within each primary area. For example, ``Economy and Jobs'' included sub-topics like ``Cost of living crisis and inflation,'' ``Housing affordability and mortgage rates,'' and ``Public sector pay and strikes'' (see \textbf{SM Section 4.5} for full list). GPT-4o then produced four distinct policy stances for each sub-topic, two liberal-leaning and two conservative-leaning. In total, this process yielded 360 issue stances. In total, these two stages yielded 744 distinct issue stances. As a final curation step, we manually reviewed and filtered these to exclude irrelevant, unclear, inappropriate, or awkwardly phrased issues.

This resulted in a final refined set of 697 uniformly phrased issue stances, covering a variety of issue areas, which we used in Study~2 and Study~3. For further description of our issue set, please see \textbf{SM Section 4.5}. For a full list of all issue stances and associated metadata, consult our \href{https://github.com/kobihackenburg/scaling-conversational-AI}{project repository}.

\subsection{Prompting for persuasiveness}
\label{sec:methods_prompting}

LLMs can be sensitive to minor changes to input prompts \cite{Röttger2024, Elazar2021, Sclar2023, Wang2023}. Additionally, there are a number of conversational persuasion techniques models could be instructed to employ. Therefore, to ensure the generalizability of our results, in all studies, models were randomized to a prompt using one of eight rhetorical strategies previously established by political persuasion literature (full prompt text can be found in \textbf{SM Section 4.4.2}):

\begin{enumerate}
\item \textbf{Information}: Focuses on presenting lots of high-quality facts, evidence, and information \cite{Petty1986, Coppock2023}.

\item \textbf{Deep canvassing}: Focuses first on comprehensively eliciting or listening to the users' views, before providing arguments \cite{Brookman2016, santoro_broockman_kalla_porat_2024}.

\item \textbf{Storytelling}: Focuses on sharing personal experiences and building compelling narratives \cite{KALLA_BROOCKMAN_2020, HAMBY201711, green2000role}.

\item \textbf{Norms}: Focuses on demonstrating that others (especially similar or important others) agree with the issue stance \cite{CIALDINI1991201, Coppock2023}.

\item \textbf{Moral re-framing}: Aligns support for the issue stance with the target audience's core moral values \cite{kalla2022personalizing, voelkel2018morally, Feinberg2019}.

\item \textbf{Debate}: Draws on a combination of distinct rhetorical elements collated via examination of transcripts of political debates in the UK House of Commons and Lords \cite{Blumenau2024}.

\item \textbf{Mega}: Model is given descriptions of all of the above strategies, can adaptively choose to use any or none.

\item \textbf{None}: Model is given no particular strategies, and is simply told to ``be as persuasive as possible''.
\end{enumerate}

\subsection{Personalization}
\label{sec:methods_personalization}

We tested personalization using three distinct methods, each intended to enhance the model's ability to tailor persuasion to individual users.

\begin{enumerate}
\item \textbf{Prompt-based personalization (Study~1):} In Study~1, we applied a simple prompt-based personalization approach. For participants assigned to the personalized condition, we appended to each model prompt (a) the participant's initial attitude score (0--100 scale), and (b) their open-ended reflection explaining their initial attitude.

\item \textbf{Fine-tuning on personalized data (Study~2):} In Study~2, we fine-tuned models using a mixed dataset in which $\sim$50\% of training conversations included personalized information. In these personalized cases, models received participants' initial attitudes and free-text justifications as well as participants' demographic and political information (age, gender, education, ideology, party affiliation, political knowledge, AI trust, attitude confidence, and issue importance).

\item \textbf{Personalized reward modeling (Studies~2 and~3):} In Studies~2 and~3, we trained a RM on data where $\sim$50\% of conversations included personalized context. For these personalized cases, models received initial attitudes and free-text justifications as well as participants' demographic and political information (age, gender, education, ideology, party affiliation, political knowledge, AI trust, attitude confidence, and issue importance). During inference, we randomized whether the RM received personalization information (50\% chance). This allowed us to assess whether incorporating richer personalization data improved the RM's ability to select persuasive responses.
\end{enumerate}

\subsection{Experiment design}
\label{sec:methods_experiment_design}
Studies 1--3 were all randomized survey experiments following a common design. Participants first passed a short writing screener, read a consent form that explained the study may involve conversation with an AI model, supplied core demographics, and were randomly assigned a single contemporary political issue. They then completed an identical three-item baseline attitude scale for that issue (measured on a 0--100 scale) and were asked to explain their attitude in an open-ended text box.

In all studies, participants were variously randomized with respect to (\textit{i}) whether the interaction took the form of a multi-turn dialogue or exposure to a static, LLM-generated message, (\textit{ii}) the specific LLM family, LLM, or post-training type, (\textit{iii}) whether the LLM employed personalization (message generated with vs.\ without the participant's personal data), and (\textit{iv}) one of eight predefined rhetorical persuasion strategies. After engaging with their assigned treatment, all participants immediately repeated the same issue attitude scale, provided an open-text rationale for any shift in attitudes between pre- and post-treatment, responded to a series of rating questions about their conversation, and received a debrief. Full allocation probabilities for each study, LLM specifications, and all question wordings are in \textbf{SM Section 3.1}.

\subsection{Statistical Analysis}
\label{sec:methods_statistical_analysis}

For ease of interpretation, we describe our statistical analyses here broadly following the subsection format of the Results section. Unless stated otherwise, to estimate treatment effects and other comparisons we use OLS with robust (HC2) standard errors \cite{mackinnon1985hc2} and adjust for participants' pre-treatment attitudes to increase precision \cite{Gerber2012}.

\paragraph{Conversation vs.\ static message and persuasion durability.} To compare the effect of AI-driven conversation and static messaging in studies~1 and~3, we exclude the control group and compute the difference in mean post-treatment attitudes directly between these conditions in both studies. In study~3 we restrict this comparison to the GPT-4.5 base model conversations, excluding the model which received our RM post-training (this ensures a fair comparison, since it was the GPT-4.5 base model that generated the static messages). To estimate the durability of the persuasive effects in study~1, to ensure a meaningful comparison, we restrict the sample of participants to those who were assigned to GPT-4o and who had non-missing outcomes both in the original study and the 1-month follow-up. The estimates are in \textbf{SM Section 2.2}.

\paragraph{Persuasive returns to model scale.} We follow our pre-registered analysis protocol to estimate the association between LLM scale and persuasiveness, which comprises three key steps. First, in each study we estimate the average treatment effect of each LLM's political conversations relative to the control group, restricting to base LLMs only—that is, excluding study conditions where LLMs received RM or SFT. Second, we pool across studies and regress these estimates onto a variable for LLM scale using robust Bayesian meta-regression with study fixed effects \cite{Hewitt2024}. We operationalize LLM scale as the logarithm (base~10) of its ``effective compute'', given by the number of floating point operations (FLOPs) \cite{Kaplan2020}. Third, to account for the fact that the association between LLM scale and persuasiveness may be either linear or nonlinear, we fit two meta-regressions; one that assumes a linear association and one that flexibly allows for a nonlinear association via fitting a generalized additive model (GAM) \cite{clarkGeneralizedAdditiveModels2022}. We then compare their out-of-sample predictive accuracies by comparing their expected log pointwise predictive densities (ELPD), estimated via leave-one-out cross-validation \cite{vehtariPracticalBayesianModel2017, burknerBrmsBookApplied2024}. We repeat this analysis three times: once for our joint scaling curve analysis that includes both chat-tuned and developer post-trained LLMs (pre-registered), and then twice more: among chat-tuned LLMs and developer post-trained LLMs separately (not pre-registered). In all cases a linear association is preferred because the GAM does not show significantly greater predictive accuracy. To estimate the interaction between LLM post-training type (chat-tuned or developer) and the scaling curve, we fit a fourth meta-regression in which we interact the linear term for the logarithm of effective compute with a dummy variable for post-training type. We summarize all estimates via the mean and 95\% percentiles of the posterior distribution. Full tables of results and diagnostics are in \textbf{SM Section 2.3.1}.

\paragraph{Persuasive returns to model post-training.} To estimate the persuasive effects of our PPT strategies, we compared them to the control group in the corresponding studies (2~and~3). To estimate the main effects of SFT and RM, and their interaction, we excluded the control group and estimated the difference in mean post-treatment attitudes directly between our different PPT conditions. Finally, to compute the average effect of RM across the developer post-trained models in studies~2 and~3, we first fitted study-level regressions with a dummy variable for RM (vs.\ not), restricting to developer post-trained models only and excluding the control group, and then we averaged these estimates weighting by their precision. Full tables of results are in \textbf{SM Section 2.4}.

\paragraph{Examining how the models persuade.} To estimate the effects of personalization, in all studies we restricted our sample to the treatment-dialogue conditions and created a dummy variable for personalization (vs.\ no personalization). We then fitted separate regressions for each unique combination of study-chat (S1 chat1, S1 chat2, S2, S3), LLM type (chat-tuned or developer post-trained) and PPT type (base, SFT, RM, SFT+RM). The overall effect was then calculated via precision-weighted averaging of these estimates (see \textbf{SM Section 2.5}). 

To estimate the effects of each prompt vs.\ the basic prompt, in all studies we restricted our sample to the treatment-dialogue conditions and created a dummy variable for each prompt (vs.\ basic prompt). We then fitted separate regressions for each unique combination of study-chat (S1 chat1, S1 chat2, S2, S3), and the overall effect was calculated via precision-weighted averaging of these estimates across study-chats. To estimate the absolute average treatment effect of the prompts (including the basic prompt), we repeated this approach but compared each prompt (including basic) to the control condition. Full tables of results are in \textbf{SM Section 2.6.1}.

We estimated the correlation and slope between information density and persuasion in two (pre-registered) steps. First, we restricted our sample to the treatment-dialogue conditions and then for each study-chat we grouped by prompt to estimate the mean number of claims (information density) made by the LLM as well as participants' post-treatment attitude (persuasion) at the prompt-level. We do this at the prompt level because prompts were randomly assigned, thereby providing exogenous variation in both information density and persuasion. Second, we then fitted two Bayesian meta-regressions on these estimates, pooling across study-chats (with fixed effects for study-chats), to estimate both the correlation and slope between information density and persuasion. This lets us account for the uncertainty in the prompt-level estimates, thus appropriately ``disattenuating'' the correlation and slope estimates. See \textbf{SM Section 2.6.1} for the meta-regression outputs and Bayesian model diagnostics.

The estimates in \textbf{\autoref{fig:fig4}C} were obtained by estimating the average treatment effect of the LLMs against the control group separately for conditions where LLMs received (\textit{i}) the information-prompt or (\textit{ii}) any other prompt. Notably, the difference in persuasion between (\textit{i}) and (\textit{ii}) is greater for both GPT-4.5 and GPT-4o (3/25) than for GPT-4o (8/24)---shown by significant ($p < .05$) interaction effects---indicating that our most persuasive models received a disproportionate increase in persuasion (vs.\ another frontier model) when prompted to deploy information (full tables of results are in \textbf{SM Section 2.6.2}.). 

The estimates in \textbf{~\autoref{fig:fig4}D} were obtained by estimating the average information density ($N$ claims) for each of the (\textit{i}) and (\textit{ii}) prompt subgroups and LLMs shown. Once again the difference in $N$ claims between (\textit{i}) and (\textit{ii}) is significantly greater for both GPT-4.5 and GPT-4o (3/25) than for GPT-4o (8/24)---indicating that our most persuasive models received a disproportionate increase in information density (vs.\ another frontier model) when prompted to deploy information (interaction effects $p < .001$, full tables of results are in \textbf{SM Section 2.6.2}). The aforementioned interaction tests were pre-registered. 

Finally, we computed the main effects of RM and SFT (\textbf{\autoref{fig:fig4}E}) on both persuasion and information density by fitting a regression on the corresponding outcome variable (post-treatment attitudes or $N$ claims) separately for studies~2 and~3 and LLM type (chat-tuned or developer), with dummy variables for RM and SFT (full tables of results are in \textbf{SM Section 2.4}).

To estimate the overall strength of association between information density and persuasion, we conducted a cross-fit, two-stage regression analysis. In the first stage, a random forest was fit to estimate the average information density in each randomized condition (based on study, model, post-training method, prompt and personalization). In the second stage, we then used the random forest model's out-of-fold predictions as input into a linear regression model to predict post-treatment attitudes (including terms for pre-treatment attitudes and study). Finally, to provide an estimate of variance explained, we compare the $R^2$ of this linear model to (\textit{a}) a baseline regression that does \textit{not} include predicted information density, and (\textit{b}) an ``upper-bound'' regression that additionally includes predictions from a random forest fit directly to predict mean attitude change by condition.

\paragraph{Examining the accuracy of the information provided by the models.} We estimated the average accuracy of claims made by individual LLMs (\textbf{\autoref{fig:fig4}A}) in two steps. First, for each participant-LLM conversation, we calculated the proportion of fact-checkable claims that were rated $> 50/100$ on the accuracy scale. Second, for each study-chat, we restricted to treatment-dialogues by base LLMs only, and then computed the mean proportion score for each LLM. Notably, this procedure excludes conversations where there were zero claims made. As described in the main text, all of the results we describe are substantively identical if we instead analyze the average accuracy score on the 0--100 scale (see \textbf{SM Section 2.7.1}). 

The estimates in (\textbf{\autoref{fig:fig4}B}) were obtained by estimating the average proportion of accurate claims separately for conditions where LLMs received (\textit{i}) the information-prompt or (\textit{ii}) any other prompt. The difference in accuracy between (\textit{i}) and (\textit{ii}) is greater for both GPT-4.5 and GPT-4o (3/25) than for GPT-4o (8/24)---shown by significant ($p < .001$) interaction effects---indicating that our most persuasive models saw a disproportionate decrease in claim accuracy (vs.\ another frontier model) when prompted to deploy information (full tables of results are in \textbf{SM Section 2.6.2}). 

We computed the main effects of RM and SFT (\textbf{\autoref{fig:fig4}C}) on both persuasion and accuracy by fitting a regression on the corresponding outcome variable (post-treatment attitudes or conversation-level accuracy score) separately for chat-tuned models in study~2, with dummy variables for RM and SFT (full tables of results are in \textbf{SM Section 2.4}).

To further test whether inaccurate claims are a byproduct or cause of increased persuasion, we performed an OLS regression that estimated the average attitude change for every randomized condition in our design, as a linear function of both the average number of inaccurate claims and total claims. In no study did we find a significant positive coefficient on inaccurate claims when adjusting for total claims (see \textbf{SM Section 2.7.2}).

Finally, to estimate the impact of a conversational model designed for maximal persuasion across all randomized features in our studies, we follow a cross-fit machine learning approach similar to that used in our analysis of information density (described above). First, a random forest was fit to estimate the average attitude change in each randomized condition (based on study, model, post-training method, prompt and personalization). We then used the random forest model's out-of-fold predictions to identify the 500 AI-conversations expected to be most persuasive across our entire dataset (excluding Study 1 conversation 1, which used a different set of issues). We report the observed average treatment effect of these 500 conversations.

\subsection{Fact-checking}
\label{sec:methods_fact-checking}

\paragraph{Fact Extraction.} We used GPT-4o to extract fact-checkable claims from each individual LLM message across all treatment-conversations in our data. In total, this resulted in $N = 466{,}769$ fact-checkable claims extracted from $N = 668{,}823$ unique LLM messages. For the prompt used for fact extraction, see \textbf{SM Section 4.4.4}.

\paragraph{Fact-checking.} Subsequently, we used a search-enabled version of OpenAI's GPT-4o model (gpt-4o-search-preview) to fact-check each claim. We instructed our fact-checking model to rate the veracity of each claim on a 0--100 scale, where 0~is completely inaccurate and 100~is completely accurate. The model was also asked to offer a brief justification for its score and provide links to any sources it used. All facts were checked with \texttt{search\_context\_size} set to \texttt{high}. The LLM fact-checking pipeline was implemented between April~1st and May~18th, 2025. For the full prompt used for fact-checking, see \textbf{SM Section 4.4.4}.

\paragraph{Validation.} To validate our AI fact-checking pipeline, we hired 2~professional fact-checkers from the \href{https://ksjfactcheck.org/}{KSJ fact-checking project} and the marketplace \href{https://www.upwork.com/}{Upwork}, and tasked them with evaluating a stratified sample of 198~LLM messages. The messages were from Study~1 Chat~2 (i.e., GPT-4o 8/24) and were stratified by the (\textit{i})~number of claims they contained and (\textit{ii})~the average accuracy of those claims, such that (\textit{i}) and (\textit{ii}) were evenly spaced from 0--10 and 0--100 respectively. 

For each of the 198~messages, we asked the fact-checkers to count both the number of fact-checkable claims contained within it, and to assign a 0--100 accuracy score to each of the resulting claims and message overall. To estimate the correlation between fact-checker and LLM ratings, for each message we averaged the fact-checker scores and then calculated the human-LLM correlation across the 198~messages---separately for both the number of claims as well as the average accuracy (see \textbf{SM Section 2.8} for break downs at the individual fact-checker level).

\begin{ack}
The authors acknowledge the use of resources provided by the Isambard-AI National AI Research Resource (AIRR). Isambard-AI is operated by the University of Bristol and is funded by the UK Government’s Department for Science, Innovation and Technology (DSIT) via UK Research and Innovation; and the Science and Technology Facilities Council [ST/AIRR/I-A-I/1023].

For help during data collection, we thank Lorna Evans, Michelle Lee, Simon Jones, and Ana Price from Prolific.  

B.M.T. acknowledges support from the Leverhulme Trust. 

\end{ack}

\section*{Author Contributions}
Conceptualization: K.H., B.M.T., D.G.R. \& C.S. \\
Experiment Design: K.H., B.M.T., L.H., D.G.R. \& C.S. \\
Model Training: K.H., L.H., \& S.B. \\
Model Hosting: E.S. \& H.L. \\
Data Analysis: K.H. \ B.M.T. \& L.H.\\
Visualization: K.H. \& B.M.T.\\
Project Support: C.F.\\
Writing - Original Draft: K.H. \& B.M.T.\\
Writing - Review and Editing: K.H., B.M.T., L.H., H.L., H.M., D.G.R. \& C.S.\\

\bibliographystyle{plain}
\bibliography{references}

\begin{thebibliography}{10}

\bibitem{altayScalingInteractiveArgumentation2022}
Sacha Altay, Marl{\`e}ne Schwartz, Anne-Sophie Hacquin, Aur{\'e}lien Allard, Stefaan Blancke, and Hugo Mercier.
\newblock Scaling up interactive argumentation by providing counterarguments with a chatbot.
\newblock {\em Nature Human Behaviour}, 6(4):579--592, February 2022.

\bibitem{argyle2025testing}
L.~P. Argyle, E.~C. Busby, J.~R. Gubler, A.~Lyman, J.~Olcott, J.~Pond, and D.~Wingate.
\newblock Testing theories of political persuasion using {AI}.
\newblock {\em Proceedings of the National Academy of Sciences}, 122(18):e2412815122, 2025.

\bibitem{bengioManagingExtremeAI2024}
Yoshua Bengio, Geoffrey Hinton, Andrew Yao, Dawn Song, Pieter Abbeel, Trevor Darrell, Yuval~Noah Harari, Ya-Qin Zhang, Lan Xue, Shai {Shalev-Shwartz}, Gillian Hadfield, Jeff Clune, Tegan Maharaj, Frank Hutter, At{\i}l{\i}m~G{\"u}ne{\c s} Baydin, Sheila McIlraith, Qiqi Gao, Ashwin Acharya, David Krueger, Anca Dragan, Philip Torr, Stuart Russell, Daniel Kahneman, Jan Brauner, and S{\"o}ren Mindermann.
\newblock Managing extreme {{AI}} risks amid rapid progress.
\newblock {\em Science}, 384(6698):842--845, May 2024.

\bibitem{Blumenau2024}
Jack Blumenau and Benjamin~E. Lauderdale.
\newblock The variable persuasiveness of political rhetoric.
\newblock {\em American Journal of Political Science}, 68:255--270, 1 2024.

\bibitem{Brookman2016}
David Broockman and Joshua Kalla.
\newblock Durably reducing transphobia: A field experiment on door-to-door canvassing.
\newblock {\em Science}, 352(6282):220--224, 2016.

\bibitem{burknerBrmsBookApplied2024}
Paul-Christian B{\"u}rkner.
\newblock {\em The Brms {{Book}}: {{Applied Bayesian Regression Modelling Using R}} and {{Stan}} ({{Early Draft}})}.
\newblock Unknown, 2024.

\bibitem{burtellArtificialInfluenceAnalysis2023}
Matthew Burtell and Thomas Woodside.
\newblock Artificial {{Influence}}: {{An Analysis Of AI-Driven Persuasion}}, March 2023.

\bibitem{chen2025frameworkassesspersuasionrisks}
Zhongren Chen, Joshua Kalla, Quan Le, Shinpei Nakamura-Sakai, Jasjeet Sekhon, and Ruixiao Wang.
\newblock A framework to assess the persuasion risks large language model chatbots pose to democratic societies, 2025.

\bibitem{CIALDINI1991201}
Robert~B. Cialdini, Carl~A. Kallgren, and Raymond~R. Reno.
\newblock A focus theory of normative conduct: A theoretical refinement and reevaluation of the role of norms in human behavior.
\newblock {\em Advances in Experimental Social Psychology}, 24:201--234, 1991.

\bibitem{clarkGeneralizedAdditiveModels2022}
Michael Clark.
\newblock {\em Generalized {{Additive Models}}}.
\newblock 2022.

\bibitem{Coppock2023}
Alexander Coppock.
\newblock {\em Persuasion in Parallel. How Information Changes Minds about Politics.}
\newblock University of Chicago Press, 2023.

\bibitem{Costello2024}
Thomas~H. Costello, Gordon Pennycook, and David Rand.
\newblock Durably reducing conspiracy beliefs through dialogues with ai.
\newblock {\em Science}, 2024.

\bibitem{delli1996politicalknowledge}
Michael~X. Delli~Carpini and Scott Keeter.
\newblock {\em What Americans Know about Politics and Why It Matters}.
\newblock Yale University Press, New Haven, CT, 1996.

\bibitem{ding2023enhancing}
Ning Ding, Yulin Chen, Bokai Xu, Yujia Qin, Zhi Zheng, Shengding Hu, Zhiyuan Liu, Maosong Sun, and Bowen Zhou.
\newblock Enhancing chat language models by scaling high-quality instructional conversations, 2023.

\bibitem{durmus2024persuasion}
Esin Durmus, Liane Lovitt, Alex Tamkin, Stuart Ritchie, Jack Clark, and Deep Ganguli.
\newblock Measuring the persuasiveness of language models, 2024.

\bibitem{El-Sayed2024}
Seliem El-Sayed, Canfer Akbulut, Amanda Mccroskery, Geoff Keeling, Zachary Kenton, Zaria Jalan, Nahema Marchal, Arianna Manzini, Toby Shevlane, Shannon Vallor, Daniel Susser, Matija Franklin, Sophie Bridgers, Harry Law, Matthew Rahtz, Murray Shanahan, Michael~Henry Tessler, Arthur Douillard, Tom Everitt, Sasha Brown, and Google Deepmind.
\newblock A mechanism-based approach to mitigating harms from persuasive generative ai.
\newblock {\em arXiv}, 2024.

\bibitem{Elazar2021}
Yanai Elazar, Nora Kassner, Shauli Ravfogel, Abhilasha Ravichander, Eduard Hovy, Hinrich Schütze, and Yoav Goldberg.
\newblock Measuring and improving consistency in pretrained language models.
\newblock {\em Transactions of the Association for Computational Linguistics}, 9:1012--1031, 9 2021.

\bibitem{epoch2023aitrends}
{Epoch AI}.
\newblock Key trends and figures in machine learning, 2023.
\newblock Accessed: 2025-06-21.

\bibitem{Feinberg2019}
Matthew Feinberg and Robb Willer.
\newblock Moral reframing: A technique for effective and persuasive communication across political divides.
\newblock {\em Social and Personality Psychology Compass}, 13, 12 2019.

\bibitem{Gerber2012}
A~S Gerber and D~P Green.
\newblock {\em Field Experiments: Design, Analysis, and Interpretation}.
\newblock W. W. Norton, 2012.

\bibitem{Goldstein2024}
Josh~A. Goldstein, Jason Chao, Shelby Grossman, Alex Stamos, and Michael Tomz.
\newblock How persuasive is ai-generated propaganda?
\newblock {\em PNAS Nexus}, 3, 2 2024.

\bibitem{Goldstein2023}
Josh~A. Goldstein, Girish Sastry, Micah Musser, Renee DiResta, Matthew Gentzel, and Katerina Sedova.
\newblock Generative language models and automated influence operations: Emerging threats and potential mitigations.
\newblock {\em arXiv}, 1 2023.

\bibitem{Grace2024}
Katja Grace, Harlan Stewart, Julia~Fabienne Sandkühler, Stephen Thomas, Ben Weinstein-Raun, and Jan Brauner.
\newblock Thousands of ai authors on the future of ai.
\newblock {\em arXiv}, 2024.

\bibitem{green2000role}
Melanie~C Green and Timothy~C Brock.
\newblock The role of transportation in the persuasiveness of public narratives.
\newblock {\em Journal of Personality and Social Psychology}, 79(5):701--721, 2000.

\bibitem{Hackenburg2024}
Kobi Hackenburg and Helen Margetts.
\newblock Evaluating the persuasive influence of political microtargeting with large language models.
\newblock {\em Proceedings of the National Academy of Sciences}, 121:e2403116121, 6 2024.

\bibitem{hackenburg2024evidence}
Kobi Hackenburg, Ben~M. Tappin, Paul Röttger, Scott Hale, Jonathan Bright, and Helen Margetts.
\newblock Scaling language model size yields diminishing returns for single-message political persuasion, 2024.

\bibitem{HAMBY201711}
Anne Hamby, David Brinberg, and Kim Daniloski.
\newblock Reflecting on the journey: Mechanisms in narrative persuasion.
\newblock {\em Journal of Consumer Psychology}, 27(1):11--22, 2017.

\bibitem{Hewitt2024}
Luke Hewitt, David Broockman, Alexander Coppock, Ben~M. Tappin, James Slezak, Valerie Coffman, Nathaniel Lubin, and Mohammad Hamidian.
\newblock How experiments help campaigns persuade voters: Evidence from a large archive of campaigns’ own experiments.
\newblock {\em American Political Science Review}, pages 1--19, 2024.

\bibitem{hsuDisinformationResearchersRaise2023}
Tiffany Hsu and Stuart~A. Thompson.
\newblock Disinformation {{Researchers Raise Alarms About A}}.{{I}}. {{Chatbots}}.
\newblock {\em The New York Times}, February 2023.

\bibitem{Ipsos2023}
Ipsos.
\newblock Global views on a.i. and disinformation, 2023.

\bibitem{jonesLiesDamnedLies2024}
Cameron~R. Jones and Benjamin~K. Bergen.
\newblock Lies, {{Damned Lies}}, and {{Distributional Language Statistics}}: {{Persuasion}} and {{Deception}} with {{Large Language Models}}, December 2024.

\bibitem{KALLA_BROOCKMAN_2020}
Joshua~L. Kalla and David~E. Brookman.
\newblock Reducing exclusionary attitudes through interpersonal conversation: Evidence from three field experiments.
\newblock {\em American Political Science Review}, 114(2):410–425, 2020.

\bibitem{kalla2022personalizing}
Joshua~L Kalla, Adam~Seth Levine, and David~E Broockman.
\newblock Personalizing moral reframing in interpersonal conversation: A field experiment.
\newblock {\em The Journal of Politics}, 84(2), 2022.
\newblock Sponsored by the Southern Political Science Association.

\bibitem{Kaplan2020}
Jared Kaplan, Sam McCandlish, Tom Henighan, Tom~B Brown, Benjamin Chess, Rewon Child, Scott Gray, Alec Radford, Jeffrey Wu, and Dario Amodei.
\newblock Scaling laws for neural language models.
\newblock {\em arXiv}, 1 2020.

\bibitem{lewkowycz2022solving}
Aitor Lewkowycz, Anders Andreassen, David Dohan, Ethan Dyer, Henryk Michalewski, Vinay Ramasesh, Ambrose Slind, Cem Anil, Imanol Schlag, Theo Gutman-Solo, et~al.
\newblock Solving quantitative reasoning problems with language models.
\newblock {\em Advances in Neural Information Processing Systems}, 35:3843--3857, 2022.

\bibitem{lucianoHypersuasionAIsPersuasive2024}
Floridi Luciano.
\newblock Hypersuasion -- {{On AI}}'s {{Persuasive Power}} and {{How}} to {{Deal}} with {{It}}.
\newblock {\em Philosophy \& Technology}, 37(2):64, May 2024.

\bibitem{Mackenzie2023}
Lucia Mackenzie and Mark Scott.
\newblock How people view ai, disinformation and elections — in charts.
\newblock {\em Politico}, 2023.

\bibitem{mackinnon1985hc2}
James~G. MacKinnon and Halbert White.
\newblock Some heteroskedasticity-consistent covariance matrix estimators with improved finite sample properties.
\newblock {\em Journal of Econometrics}, 29(3):305--325, 1985.

\bibitem{matzPotentialGenerativeAI2024}
S.~C. Matz, J.~D. Teeny, S.~S. Vaid, H.~Peters, G.~M. Harari, and M.~Cerf.
\newblock The potential of generative {{AI}} for personalized persuasion at scale.
\newblock {\em Scientific Reports}, 14(1):4692, February 2024.

\bibitem{mercierNotBornYesterday2020}
Hugo Mercier.
\newblock {\em Not {{Born Yesterday}}: {{The Science}} of {{Who We Trust}} and {{What We Believe}}}.
\newblock Princeton University Press, January 2020.

\bibitem{mercier2011argumentative}
Hugo Mercier and Dan Sperber.
\newblock Why do humans reason? arguments for an argumentative theory.
\newblock {\em Behavioral and Brain Sciences}, 34(2):57--111, 2011.

\bibitem{mercierEnigmaReason2018}
Hugo Mercier and Dan Sperber.
\newblock {\em The {{Enigma}} of {{Reason}}}.
\newblock Harvard University Press, September 2018.

\bibitem{nostaAIsSuperhumanPersuasion}
John Nosta.
\newblock {{AI}}'s {{Superhuman Persuasion}} {\textbar} {{Psychology Today}}.
\newblock https://www.psychologytoday.com/intl/blog/the-digital-self/202310/ais-superhuman-persuasion.

\bibitem{ouyang2022training}
Long Ouyang, Jeffrey Wu, Xu~Jiang, Diogo Almeida, Carroll Wainwright, Pamela Mishkin, Chong Zhang, Sandhini Agarwal, Katarina Slama, Alex Ray, et~al.
\newblock Training language models to follow instructions with human feedback.
\newblock {\em Advances in Neural Information Processing Systems}, 35:27730--27744, 2022.

\bibitem{Peer2022}
Eyal Peer, David Rothschild, Andrew Gordon, Zak Evernden, and Ekaterina Damer.
\newblock Data quality of platforms and panels for online behavioral research.
\newblock {\em Behavior Research Methods}, 54:1643, 8 2022.

\bibitem{Petty1986}
Richard~E Petty and John~T Cacioppo.
\newblock {\em The Elaboration Likelihood Model of Persuasion}, pages 1--24.
\newblock Springer New York, 1986.

\bibitem{pilz2025trendsaisupercomputers}
Konstantin~F. Pilz, James Sanders, Robi Rahman, and Lennart Heim.
\newblock Trends in ai supercomputers, 2025.

\bibitem{prior2007postbroadcast}
Markus Prior.
\newblock {\em Post-Broadcast Democracy: How Media Choice Increases Inequality in Political Involvement and Polarizes Elections}.
\newblock Cambridge University Press, Cambridge, 2007.

\bibitem{rogiersPersuasionLargeLanguage2024}
Alexander Rogiers, Sander Noels, Maarten Buyl, and Tijl~De Bie.
\newblock Persuasion with {{Large Language Models}}: A {{Survey}}, November 2024.

\bibitem{Röttger2024}
Paul Röttger, Valentin Hofmann, Valentina Pyatkin, Musashi Hinck, Hannah~Rose Kirk, Hinrich Schütze, and Dirk Hovy.
\newblock Political compass or spinning arrow? towards more meaningful evaluations for values and opinions in large language models.
\newblock {\em arXiv}, 2 2024.

\bibitem{Salvi2024}
Francesco Salvi, Manoel~Horta Ribeiro, Riccardo Gallotti, and Robert West.
\newblock On the conversational persuasiveness of large language models: A randomized controlled trial.
\newblock {\em arXiv}, 3 2024.

\bibitem{santoro_broockman_kalla_porat_2024}
Erik Santoro, David Broockman, Joshua Kalla, and Roni Porat.
\newblock Listen for a change? a longitudinal field experiment on listening's potential to facilitate persuasion, Sep 2024.

\bibitem{schoenegger2025large}
Philipp Schoenegger, Francesco Salvi, Jiacheng Liu, Xiaoli Nan, Ramit Debnath, Barbara Fasolo, Evelina Leivada, Gabriel Recchia, Fritz G{\"u}nther, Ali Zarifhonarvar, et~al.
\newblock Large language models are more persuasive than incentivized human persuaders.
\newblock {\em arXiv preprint arXiv:2505.09662}, 2025.

\bibitem{Sclar2023}
Melanie Sclar, Yejin Choi, Yulia Tsvetkov, Alane Suhr, and Paul~G Allen.
\newblock Quantifying language models' sensitivity to spurious features in prompt design or: How i learned to start worrying about prompt formatting.
\newblock {\em arXiv}, 10 2023.

\bibitem{sevilla2024scaling}
Jaime Sevilla, Tamay Besiroglu, Ben Cottier, Josh You, Edu Rold{\'a}n, Pablo Villalobos, and Ege Erdil.
\newblock Can {AI} scaling continue through 2030?
\newblock Blog post.

\bibitem{simchonPersuasiveEffectsPolitical2024}
Almog Simchon, Matthew Edwards, and Stephan Lewandowsky.
\newblock The persuasive effects of political microtargeting in the age of generative artificial intelligence.
\newblock {\em PNAS Nexus}, 3(2):pgae035, February 2024.

\bibitem{Stagnaro2024}
Michael~Nicholas Stagnaro, James Druckman, Adam~J. Berinsky, Antonio~Alonso Arechar, Robb Willer, and David Rand.
\newblock Representativeness versus response quality: Assessing nine opt-in online survey samples.
\newblock {\em OSF Preprints}, 2 2024.

\bibitem{tappinQuantifyingPotentialPersuasive2023}
Ben~M. Tappin, Chloe Wittenberg, Luke~B. Hewitt, Adam~J. Berinsky, and David~G. Rand.
\newblock Quantifying the potential persuasive returns to political microtargeting.
\newblock {\em Proceedings of the National Academy of Sciences}, 120(25):e2216261120, June 2023.

\bibitem{vehtariPracticalBayesianModel2017}
Aki Vehtari, Andrew Gelman, and Jonah Gabry.
\newblock Practical {{Bayesian}} model evaluation using leave-one-out cross-validation and {{WAIC}}.
\newblock {\em Statistics and Computing}, 27(5):1413--1432, September 2017.

\bibitem{voelkel2018morally}
Jan~G Voelkel and Matthew Feinberg.
\newblock Morally reframed arguments can affect support for political candidates.
\newblock {\em Social Psychological and Personality Science}, 9(8):917--924, 2018.

\bibitem{Wack2025}
Morgan Wack, Carl Ehrett, Darren Linvill, and Patrick Warren.
\newblock Generative propaganda: Evidence of ai’s impact from a state-backed disinformation campaign.
\newblock {\em PNAS Nexus}, 4(4):pgaf083, 04 2025.

\bibitem{Wang2023}
Jindong Wang, Xixu Hu, Wenxin Hou, Hao Chen, Runkai Zheng, Yidong Wang, Linyi Yang, Wei Ye, Haojun Huang, Xiubo Geng, Binxing Jiao, Yue Zhang, and Xing Xie.
\newblock On the robustness of chatgpt: An adversarial and out-of-distribution perspective.
\newblock {\em arXiv}, 2 2023.

\bibitem{wei2022chain}
Jason Wei, Xuezhi Wang, Dale Schuurmans, Maarten Bosma, Fei Xia, Ed~Chi, Quoc~V Le, Denny Zhou, et~al.
\newblock Chain-of-thought prompting elicits reasoning in large language models.
\newblock {\em Advances in Neural Information Processing Systems}, 35:24824--24837, 2022.

\end{thebibliography}

\end{document}